\documentclass[journal]{IEEEtran}

\IEEEoverridecommandlockouts                              

\usepackage{tikz}              
\usepackage{tikz-3dplot}
\usepackage{tikz-qtree}
\usepackage{cite}
\usepackage{algpseudocode}
\usepackage{multirow}
\usepackage{amsmath}
\usepackage{amssymb}
\usepackage{amsmath} 

\usepackage[flushleft]{threeparttable}
\usepackage{subfigure}

\usepackage{tabularx} 
\DeclareGraphicsExtensions{.pdf,.jpg,.png}

\usepackage{graphicx}
\usepackage{caption}
\usepackage{float}
\usepackage[english]{babel}
\usepackage[linesnumbered,ruled]{algorithm2e}
\addto\captionsenglish{}

\usepackage{caption}
\usepackage{float}
\title{Evaluating Scenario-based Decision-making for Interactive Autonomous Driving Using Rational Criteria: A Survey 
}

\author{Zhen Tian$^{\dagger}$, 
         Zhihao Lin$^{\dagger}$, Dezong Zhao$^{\star}$,~\IEEEmembership{Senior Member,~IEEE}, Wenjing Zhao, David Flynn,~\IEEEmembership{Member,~IEEE}, Shuja Ansari and Chongfeng Wei,~\IEEEmembership{Member,~IEEE}
\thanks{The authors are with the James Watt School of Engineering, University of Glasgow, Glasgow, G12 8QQ, U.K.}%
\thanks{$\star$ Corresponding author: dezong.zhao@glasgow.ac.uk.}
\thanks{$\dagger$ Equal contribution}
}

\begin{document}
\maketitle


\begin{abstract}
Autonomous vehicles (AVs) can significantly promote the advances in road transport mobility in terms of safety, reliability, and decarbonization. However, ensuring safety and efficiency in interactive during within dynamic and diverse environments is still a primary barrier to large-scale AV adoption. In recent years, deep reinforcement learning (DRL) has emerged as an advanced AI-based approach, enabling AVs to learn decision-making strategies adaptively from data and interactions. DRL strategies are better suited than traditional rule-based methods for handling complex, dynamic, and unpredictable driving environments due to their adaptivity. However, varying driving scenarios present distinct challenges, such as avoiding obstacles on highways and reaching specific exits at intersections, requiring different scenario-specific decision-making algorithms. Many DRL algorithms have been proposed in interactive decision-making. However, a rationale review of these DRL algorithms across various scenarios is lacking. Therefore, a comprehensive evaluation is essential to assess these algorithms from multiple perspectives, including those of vehicle users and vehicle manufacturers. This survey reviews the application of DRL algorithms in autonomous driving across typical scenarios, summarizing road features and recent advancements. The scenarios include highways, on-ramp merging, roundabouts, and unsignalized intersections. Furthermore, DRL-based algorithms are evaluated based on five rationale criteria: driving safety, driving efficiency, training efficiency, unselfishness, and interpretability (DDTUI). Each criterion of DDTUI is specifically analyzed in relation to the reviewed algorithms. Finally, the challenges for future DRL-based decision-making algorithms are summarized.

\end{abstract}

\begin{IEEEkeywords}
Interactive autonomous driving, decision making, deep reinforcement learning, typical scenarios, rationale evaluation.
\end{IEEEkeywords}

\section{INTRODUCTION}

\IEEEPARstart{A}{U}{T}{O}{N}{O}{M}{O}{U}{S} vehicles (AVs) face significant challenges in making reliable decisions when interacting with human-driven vehicles (HDVs). This challenge is primarily due to the difficulty of accurately predicting the intentions of HDVs. Road traffic crashes cause significant fatalities and serious injuries, reflecting the global issue of millions of lives lost annually \cite{UKGov2023}. Since 2021, over 900 Tesla crashes involving driver-assistance systems have been reported \cite{omeiza2021explanations}. Despite unresolved safety issues, the number of AVs is projected to surpass 50 million by 2024 \cite{ignatious2022overview}. These statistics underscore the critical need for improving safety in autonomous driving. With a safe decision-making system, AVs have the potential to significantly decrease road crashes caused by human errors such as fatigue, distraction, and delayed reactions \cite{BADUE2021113816}. Moreover, AVs are capable of making optimal decisions faster than human drivers, thereby enhancing traffic efficiency \cite{perez2011autonomous}. 
\begin{figure*}[t]
    \centering
    \includegraphics[width=0.8\linewidth]{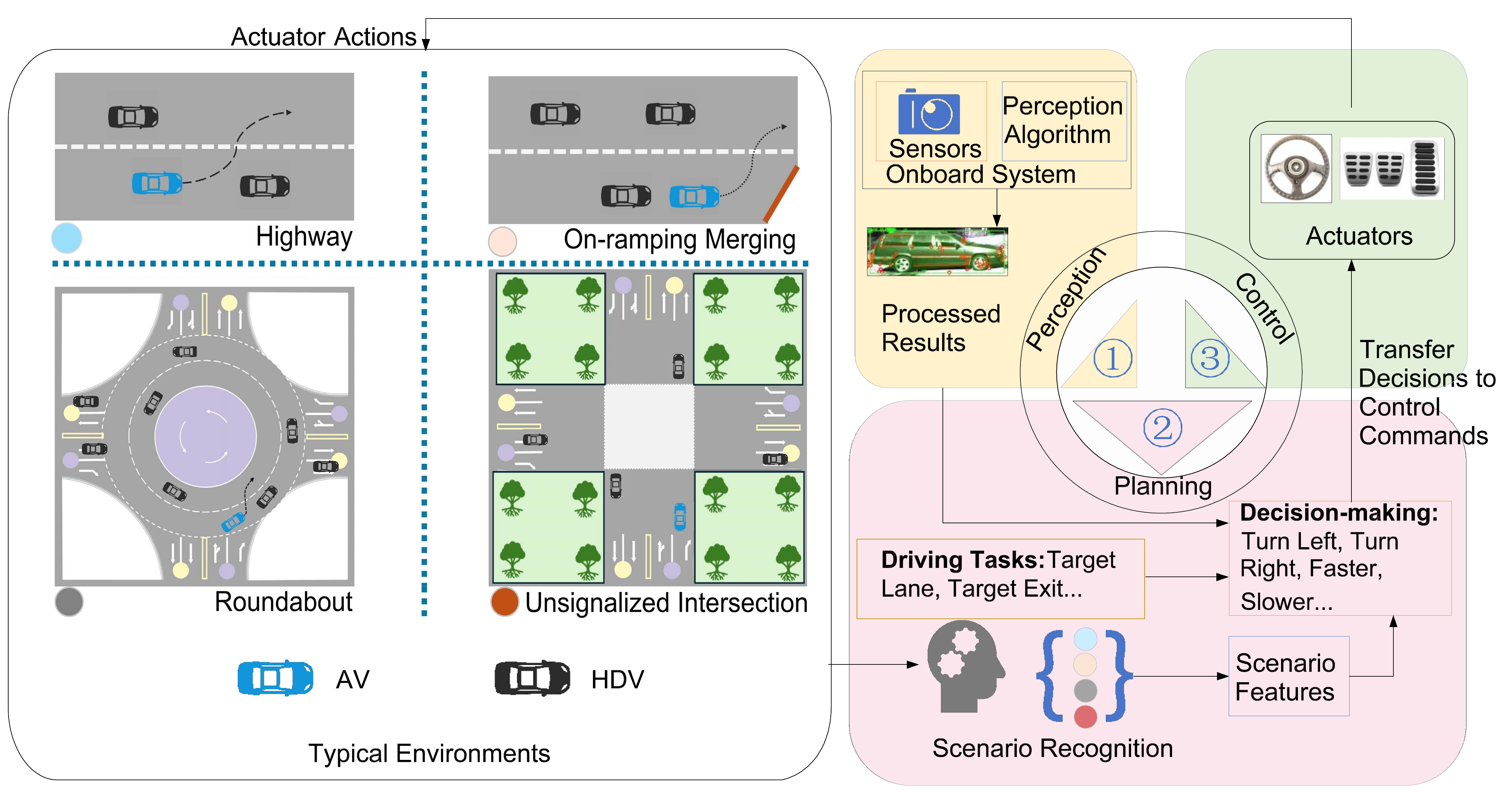}
    \caption{Autonomous driving in different scenarios.}
    \label{fig2_interactive}
\end{figure*}
\begin{figure*}[h]
    \centering
    \includegraphics[width=0.7\linewidth]{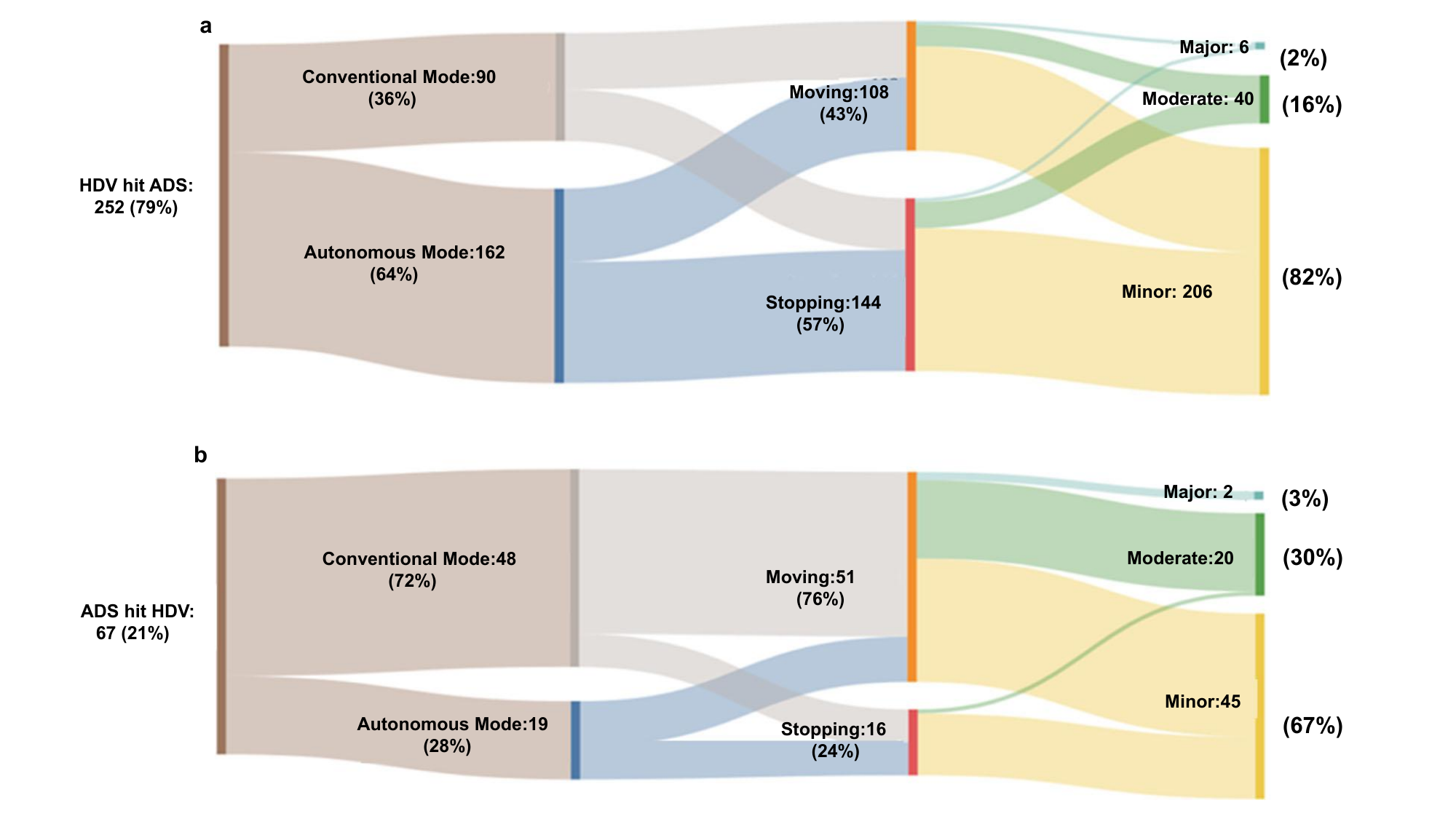}
    \caption{Rear-end accident conditions between ADS and HDV: (a) Rear-end accidents that HDV hit an ADS from behind with a sample of 252; (b) Rear-end accidents that ADS hit an HDV from behind with a sample of 67 [16].}
    \label{fig2_interactive}
\end{figure*}
There are several typical driving scenarios, such as highways, roundabouts, on-ramping merging, and unsignalized intersections, each characterized by distinct road features and scenario-specific requirements. Autonomous driving in such scenarios is depicted in Fig. 1. For example, on-ramp merging involves completing lane changes well in advance of any obstructed roadway, while navigating a roundabout requires seamlessly exiting at the intended point. Achieving these scenario-based requirements relies heavily on precise and timely operational decision-making in real time. Operational decision support for AV driving includes perception, planning, and control modules. The perception module consists of onboard sensors that continuously perceive the surrounding environment. The perceived data is processed through perception algorithms, such as YOLO methods \cite{lin2024dpl,lin2024slam2}. The planning module handles driving tasks based on scenario recognition. Subsequently, the motion planner generates discrete decisions and converts them into feasible trajectories. These feasible trajectories are then transmitted to the control module to generate control commands, which are sent to the vehicle's actuators. The actuators, including the steering wheel and pedals, receive and execute the control commands to drive the vehicle.

The interactions between AVs and HDVs are complex and therefore continuous decision-making is required, such as lane changes or braking \cite{WHO2023}.   The model-based, simple guidance, and learning-based methods are commonly used in interactive driving with HDVs.

 There are mainly four types of model-based approaches. The first model-based approach aims to predict the intentions or trajectories of HDVs, but heavily relies on rule-based classification. For example, \cite{10255298} predicts the trajectories of HDVs within a fixed time window. However, the time required for a lane-changing maneuver may exceed this fixed time window. The second model-based approach is to make decisions using robust control methods, such as the min-max model predictive control \cite{lofberg2003minimax}. However, robust control methods make excessively cautious decisions based on a worst-case scenario assumption \cite{rakovic2016model}. These methods are not suitable for most real traffic environments because worst-case scenarios are rare in real-world settings. Furthermore, decisions made for worst-case scenarios negatively impact driving quality, such as resulting in slower driving speeds. On the other hand, the game theory, the third model-based approach, has gained popularity recently. Game theory includes cooperative and non-cooperative games, both relying on equilibrium models. However, these models fail to capture the complexities of real-world driving, which are characterized by uncertainties and do not adhere to a regular equilibrium framework. Therefore, model-based methods are unable to handle interactive driving with HDVs effectively. Additionally, the fourth model-based approach, including collision-avoidance methods \cite{lin2024enhanced} and Voronoi diagram-based methods \cite{4276103}, is unable to safely respond to movable objects. Real-world collisions between HDVs and vehicles equipped with advanced driving system (ADS) assistance are summarized in \cite{abdel2024matched}. As illustrated in Fig. 2, $79~\%$ of accidents involve HDVs hitting AVs, and $21~\%$ of that involve AVs hitting HDVs. Therefore, achieving collision-free interactions with HDVs are still to be addressed.

 Compared to the aforementioned methods, simple guidance methods, such as risk-quantified fields, are widely used because they do not need to predict HDVs' intentions or make excessively cautious decisions \cite{triharminto2016novel}. The artificial potential field (APF) is a typical example, which can guide the AV to the target lane without collisions by utilizing attractive and repulsive force fields \cite{yao2020path}. However, APF assumes that all areas around the vehicle have the same level of risk because it calculates risks toward the central point. This assumption differs from reality, where the front of a car faces more danger than other parts. Additionally, APF is difficult to generalize across different scenarios without prior knowledge of the entire environment~\cite{triharminto2017local}.

To promote collision-free interactions, a large number of interactions are needed to exclude risky actions, taking into account the uncertainties in decision-making and the varying driving conditions of HDVs. Learning-based methods facilitate the exploration of control strategies by allowing full interaction with the mixed-traffic environment. These methods enable AVs to learn and adapt to complex driving scenarios through iterative interactions and feedback. Machine learning (ML) \cite{mahesh2020machine,jordan2015machine} focuses on developing algorithms to make decisions based on data, including supervised, unsupervised, and reinforcement learning. Supervised learning trains models on labeled data, supporting tasks like classification~\cite{muhammad2015supervised,castelli2018supervised}. However, supervised learning is less suited for implementation in real driving environments, as labeling complex driving scenarios exhaustively is challenging and impractical. Unsupervised learning methods are particularly suitable for interactive driving as they do not require labeled data, allowing agents to learn decision-making strategies independently. Unsupervised machine learning has demonstrated robust performance across a range of driving scenarios \cite{tian2018adaptive}. However, unsupervised learning often struggles with generalization in highly dynamic environments. Reinforcement learning (RL) is a powerful technique for making optimal decisions in dynamic environments \cite{kaelbling1996reinforcement,lu2023event}. RL involves an agent that interacts with its environment and learns safe control strategies through a reward-based framework. The adaptability of RL makes it ideal for interactive driving, where the environment is constantly changing, and the AV must adjust its behavior accordingly.
\begin{figure}[t]
    \centering
    \includegraphics[width=0.8\linewidth]{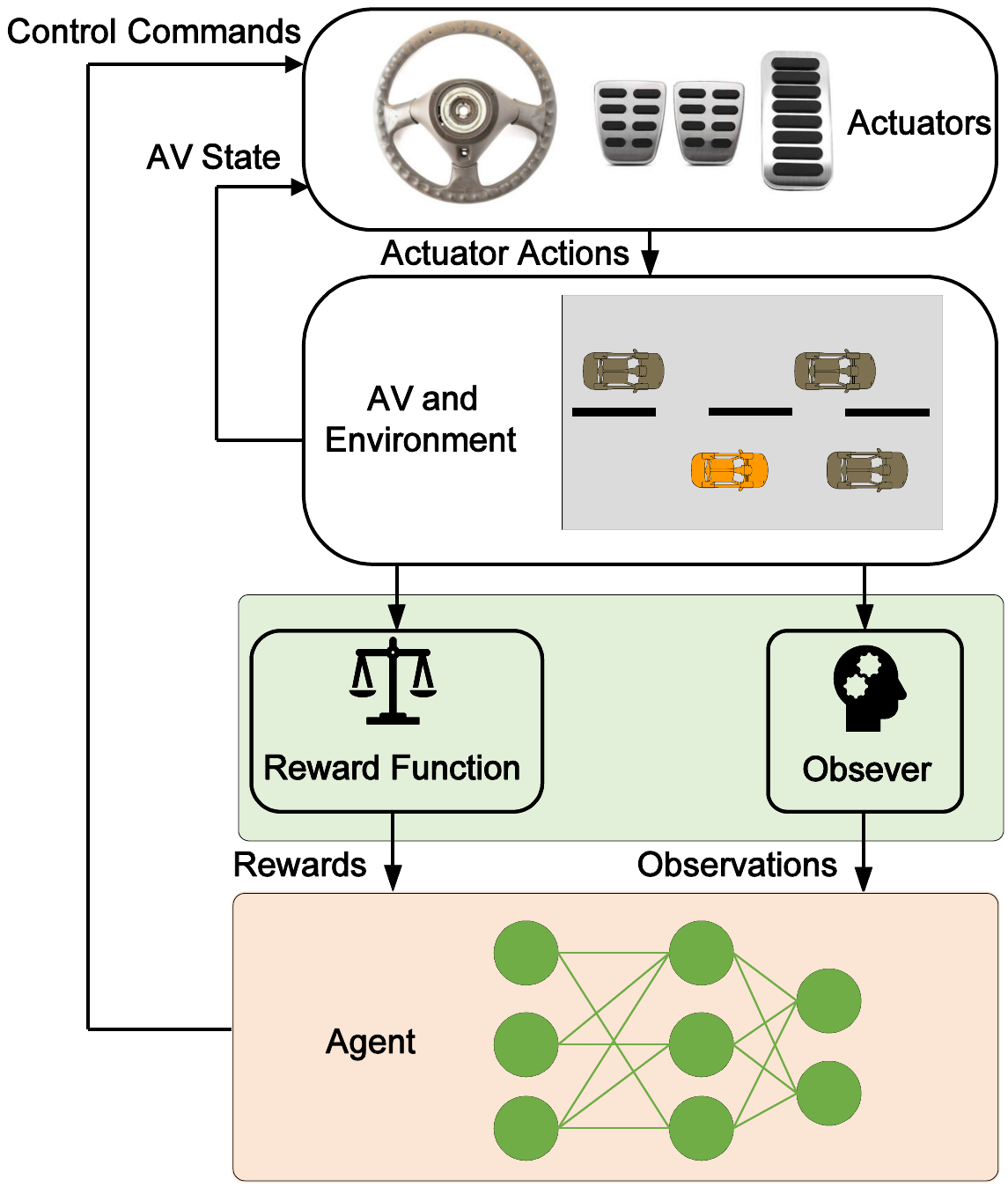}
    \caption{DRL-based autonomous driving system}
    \label{fig2_interactive}
\end{figure}

Deep reinforcement learning (DRL) is an advanced form of RL that combines the principles of deep learning \cite{lecun2015deep,goodfellow2016deep} with RL. By utilizing deep neural networks to approximate complex value functions, DRL enables agents to learn directly from perceptual inputs, such as sensory data. This capability allows DRL to handle more complex and real-time decision-making tasks compared to traditional RL. For example, \cite{yeom2022deep} demonstrates the application of DRL in collision-free path planning against surrounding obstacles.

The DRL-based autonomous driving system is illustrated in Fig. 3. The agent interacts with the environment through actuators, observations, and rewards. The agent comprises a decision network that receives information from observations of the environment and uses rewards to assess its actions. These observations are provided by the observer, which interprets the state of the AV and its environment. Based on the observations, the agent generates control commands and then sends the commands to actuators. Following the actuation of these control commands, the renewed environment information and AV state are updated. Simultaneously, a reward function evaluates the agent's actions based on predefined metrics such as safety, efficiency, or compliance to driving norms. This reward function assigns positive or negative rewards depending on how well the AV’s actions align with the desired outcomes. These rewards are then fed back to the agent, guiding the learning towards the optimal driving behavior.

\begin{figure*}[ht]
    \centering
    \includegraphics[width=0.9\linewidth]{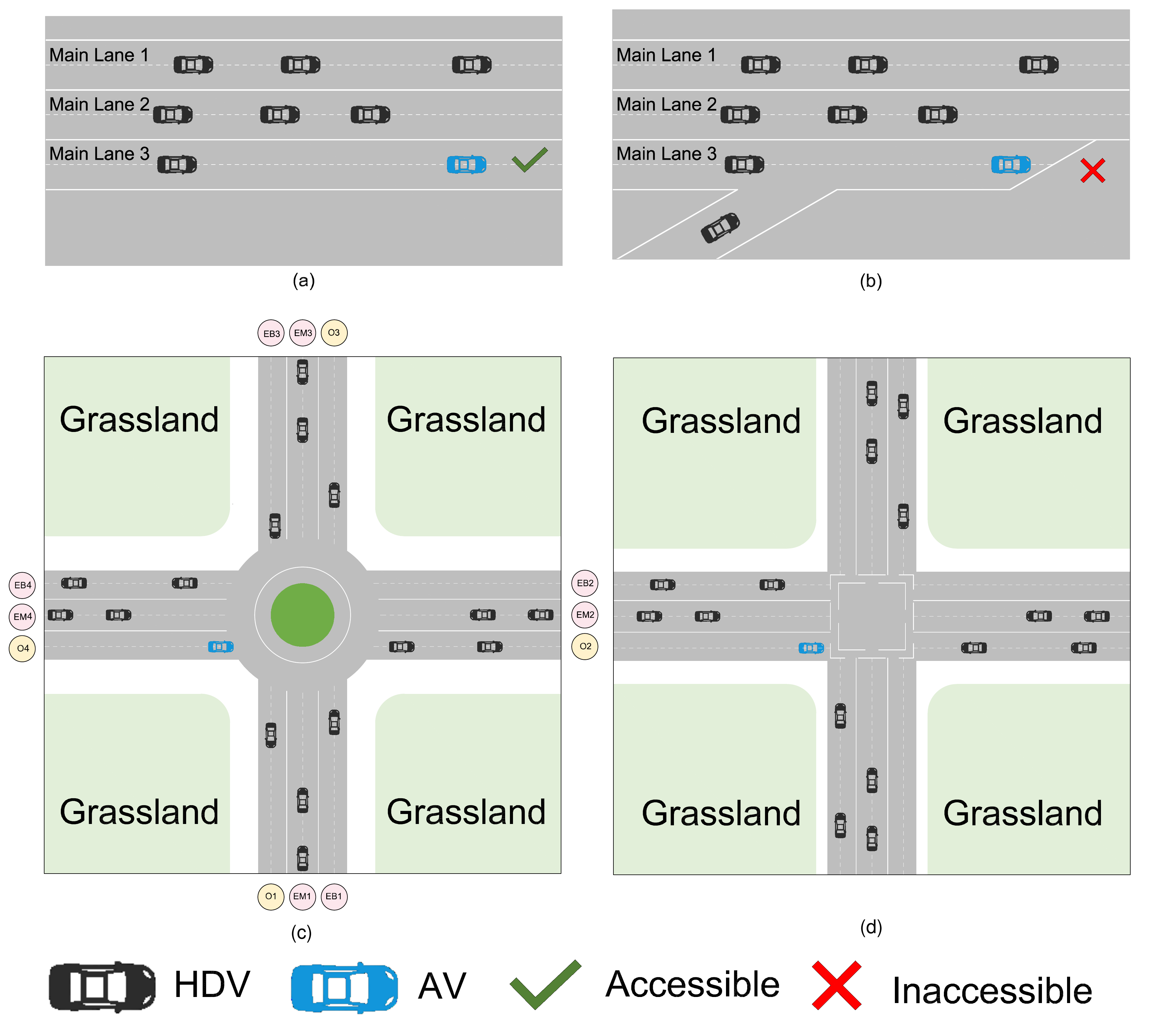}
   \caption{Example scenarios of autonomous driving: (a) highway; (b) on-ramp merging; (c) roundabout with 12 ports (8 entrances: EM1–EM4, EB1–EB4; 4 exits: O1–O4) and a central planted island; (d) unsignalized intersection.}
    \label{fig2_interactive}
\end{figure*}

DRL has been proven effective in handling emergency situations, which are critical for real-world driving scenarios. For example, \cite{muzahid2022deep} proposes a DRL-powered driving system designed to avoid collisions in emergencies. This system learns to react swiftly and safely to sudden changes, improving the robustness of decision-making in real-world conditions. Recently, several studies have been demonstrated in various scenarios \cite{xu2020nash,min2018deep,gu2016continuous,wei2019mixed,ye2020automated,dulac1512deep,tian2024efficient,tian2024balanced,basile2022ddpg,hebaish2022towards}. However, different scenarios present distinct driving requirements, necessitating tailored algorithms.
On highways, the decision-making of AVs primarily focuses on avoiding collisions with HDVs while maintaining a high average speed. In contrast, ramps introduce additional challenges, such as blocked areas that are not present on highways, as illustrated in Fig. 4. Furthermore, it is essential to assess DRL-based algorithms based on demands from various social perspectives, including vehicle users, vehicle manufacturers, and public traffic systems. Research on DRL-based algorithms, categorized by driving scenarios and evaluated based on their adaptability to real-world demands, is crucial for identifying valuable research directions.

This survey aims to review DRL-based algorithms for autonomous interactive driving, classified by scenarios and evaluated for adaptation to real-world conditions. Four typical scenarios are included: highways, on-ramping merging, roundabouts, and unsignalized intersections. Five key evaluation factors are used, including driving safety \cite{abualigah2023modified,tang2023survey}, driving efficiency \cite{10107652,wu2023deep}, training efficiency \cite{10205502,tian2024efficient}, unselfishness \cite{yue2024navigating,toghi2022social,liu2024sociality}, and interpretability \cite{huang2020survey,fan2021interpretability} (DDTUI). DRL-based decision-making approaches are reviewed for the four typical scenarios and evaluated using the criteria of DDTUI. The evaluation is consistent across all papers by examining the inclusion of evaluation factors in the designed algorithms and their corresponding verifications. For example, if a paper discusses safety but doesn’t include verifications like a lower number of collisions or consistently maintaining safe distances $d_s$, it would not be considered to include the safety factor. The reminder of this survey is structured as follows: Section II describes the features and driving tasks of the four typical scenarios. Section III explains the rationale for selecting the five evaluation factors. Sections IV-VII evaluated DRL-based algorithms for highways, on-ramping merging, roundabouts, and unsignalized intersections using the criteria of DDTUI, respectively. Finally, Section VIII provides conclusions and discussions.

\section{Road Features and Driving Tasks}
This section provides the road features and driving tasks for AVs in the scenarios of highways, on-ramping merging, roundabouts, and unsignalized intersections.

\subsection{Highways}
\subsubsection{Road Features and Driving Tasks}
Highways are fundamental components of road networks, designed to enable vehicle movement over long distances with minimal interruption. The design of highways focuses on safety, efficiency, and environmental impact. Safety features include wide lanes and clear signage to reduce collision risks.High efficiency is achieved by optimizing lane layouts to keep vehicles driving smoothly and reduce bottlenecks. The impact of highways on natural landscapes is reduced through careful route planning. The \emph{Interstate Highway System} in the United States is a vast network of highways designed to support long-distance travel and economic connectivity across states \cite{karas2015highway}. Similarly, Germany’s \emph{Autobahn}, known for its sections without speed limits, exemplifies the balance between high-speed travel and safety on highways \cite{gross2020speed}.

\subsubsection{An Example of a Highway} Fig. 4(a) presents a scenario involving a three-lane highway. The AV drives in main lane 3 and interacts with HDVs in all three lanes. There are no disturbances or uncertainties other than the surrounding HDVs. Therefore, the issue of driving safety primarily relates to collisions with surrounding HDVs. In the car-following phase, the AV can follow the HDV ahead by adjusting its acceleration. However, cautious following can lead to a loss of driving efficiency. To maintain high driving efficiency, the AV may change lanes when the space ahead is limited. However, collisions with HDVs in the target lane could occur during the lane-changing. Therefore, the driving task on highways can be summarized as balancing collision avoidance with surrounding HDVs while maintaining a consistently high speed.

\subsection{On-ramping Merging}
\subsubsection{Road Features and Driving Tasks}
Ramps, including on-ramps or off-ramps, are essential components of highway systems. Due to the symmetry between on-ramping and off-ramping processes, this survey considers only on-ramping merging. Ramps enable the smooth and safe transition of vehicles between different roadways, typically connecting surface streets with highways. Ramps provide access to highways without disrupting traffic flow on the main highway lanes.

Ramp design focuses on safety, efficiency, and space utilization. Safety is crucial, as ramps must accommodate vehicles accelerating or decelerating while merging onto or diverging from the highway. On-ramps enhance traffic flow by reducing disruptions to mainline traffic and providing sufficient space for safe  merging. Additionally, urban space constraints often require innovative ramp designs, such as cloverleaf interchanges, to connect multiple roadways effectively.
\subsubsection{Comparison with Highways}
Highways and ramps serve different functions, which are summarized below.

\begin{itemize}
    \item Functionality: Highways are designed for high-speed, long-distance travel with minimal interruption, while ramps are the transition between different road types.
    \item Design: Highways are characterized by long, straight stretches with multiple lanes, designed to maintain high speeds and efficient traffic flow. In contrast, ramps often involve curves and elevation changes, designed to accommodate vehicles as they speed up or slow down.
    \item Speed: Highways support higher speeds, with vehicles typically traveling at constant high speeds over long distances. Ramps involve acceleration or deceleration, requiring careful design to manage the speed differential between the ramp lane and the main lane.
\end{itemize}

For example, the \emph{Cloverleaf Interchange} is a common design that efficiently manages space while connecting highways with multiple surface streets \cite{leisch1993freeway}. Another example is the \emph{High Occupancy Vehicle (HOV) lane ramps}, which are designed to control the flow of carpool vehicles onto highways, providing direct and less congested access points \cite{davis2000ramp}.

Consider a three-lane ramp scenario in Fig. 4(b), which includes two main lanes and one ramp lane. The AV interacts with both dynamic and static objects. The dynamic objects are surrounding HDVs, each with unique driving intentions, speeds, and acceleration patterns. The static object represents an obstruction within the ramp lane, rendering the lane impassable and blocking access. As a result, the AV must change into the main lane before the ramp ends, considering the HDVs and the feasibility in lane-changing.

Waiting for enough space to change lanes and driving slowly to avoid blocked roads lead to safer driving. However, this cautious driving can significantly reduce driving efficiency and lower road capacity on the ramp. Consequently, it is challenging to navigate the ramp, avoid collisions with both surrounding HDVs and the blocked road ahead, while still maintaining a high driving speed.

\subsection{Roundabouts}
\subsubsection{Road Features and Driving Tasks}
Roundabouts are designed to improve traffic flow and enhance safety by reducing the likelihood of severe accidents. One example of a typical roundabout is \emph{Folon’s obelisk in Pietrasanta} in Italy, which features a central island and circular roads around it \cite{pratelli2009visibility}. Another example is the \emph{Place Charles de Gaulle} in Paris, France, where twelve major avenues converge around the Arc de Triomphe \cite{9922249}. 
\subsubsection{An Example of a Roundabout} 
An example of a roundabout is presented in Fig. 4(c). The AV starts from the EB4 port and has three possible exit choices: O1, O2, and O3. When the target exit is O1, the AV can simply follow the outer lane. For the target exit O2, there are two possible routes. One route is staying in the outer lane, which is generally safer. The other route is merging into the inner lane and exiting near O2. This second route is more efficient, as the inner lane offers a shorter curve length for the same round angle. However, rear vehicles driving in the inner lane bring potential collision risks. For the target exit O3, the AV must find the right moment to merge into the inner lane. After traveling in the inner lane for a period, the AV is expected to change lanes near the exit. The main challenge is to safely interact with other HDVs when approaching each of these three exits.
\begin{figure*}[t]
    \centering
    \includegraphics[width=0.9\linewidth]{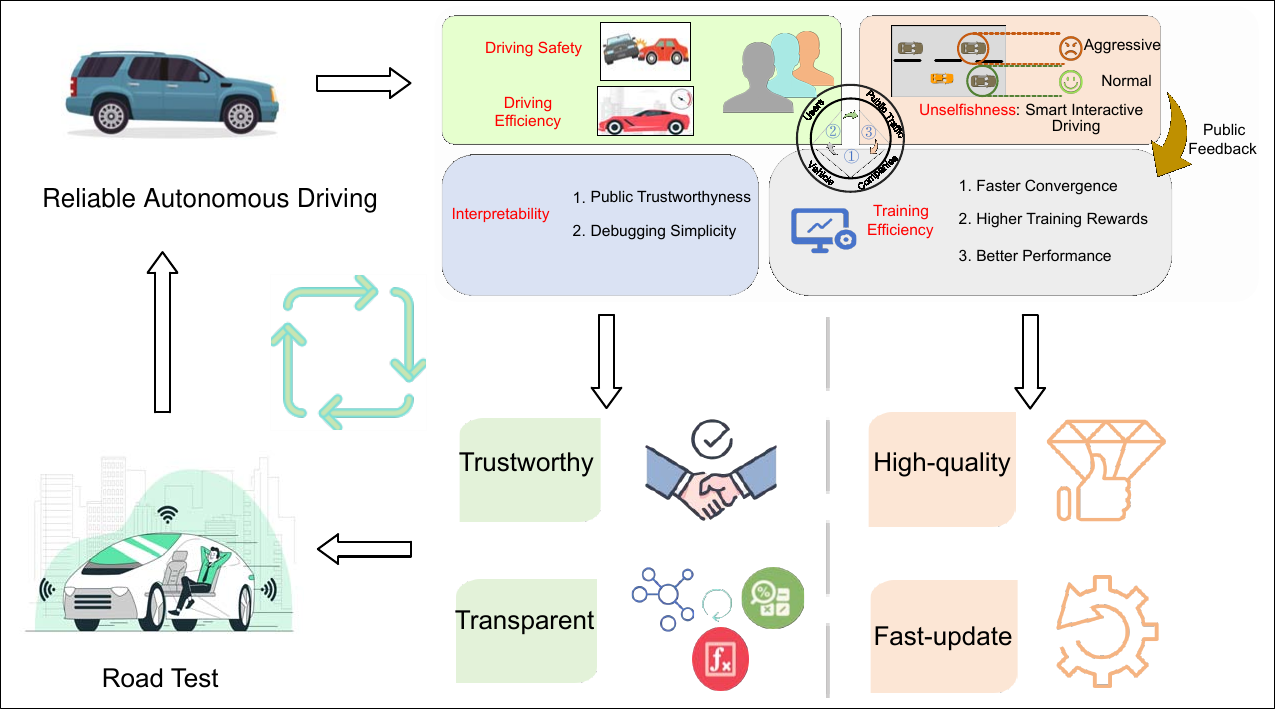}
    \caption{The importance and necessaries of achieving DDTUI in real-world autonomous driving.}
    \label{fig7}
\end{figure*}

\subsection{Unsignalized Intersections}
\subsubsection{Road Features and Driving Tasks}
Unsignalized intersections are critical components of road networks where two or roads meet or cross. They are designed to manage traffic flow from different directions, enabling vehicles to navigate safely through crossing points. Unsignalized intersection can control and organize traffic movements, reduce congestion, and enhance safety for all vehicles. One example is the \emph{Diverging Diamond Interchange (DDI)} \cite{bared2005design}.

\subsubsection{An Example of an Unsignalized Intersection} 
Fig. 4(d) shows a three-lane unsignalized intersection designed for moderate to heavy traffic flow. The intersection accommodates vehicles from all four directions, with dedicated lanes for specific traffic movements. Each approach to the intersection includes three lanes, and the areas surrounding the intersection are grassland. At the center, where all four roads meet, there is an ample space for vehicles to make turns from any direction. This central area is essential for preventing bottlenecks and ensuring smooth traffic flow.

\section{Rationale of the Evaluation Factors}
In the context of adapting decision-making algorithms to real-world driving, five key evaluation factors have been selected: driving safety and efficiency, training efficiency, unselfishness, and interpretability. 
As depicted in Fig. 5, driving safety and efficiency form the foundation of any autonomous driving system. Training efficiency enables faster convergence of algorithms. Unselfishness enhances interaction with surrounding traffic, promoting cooperation with HDVs. Meanwhile, interpretability fosters public trust and addresses algorithmic errors, ensuring that decision-making is transparent and understandable. The detailed rationale behind selecting these factors is discussed below.

\subsection{Driving Safety}
Driving safety is a fundamental requirement for autonomous vehicles. Frequent collisions cause substantial economic losses and pose severe safety risks \cite{york1997economic,burnett2023traffic}. Therefore, driving safety is primarily evaluated based on the frequency of collisions with other vehicles \cite{wang2016driving,nahata2021assessing}. Minimizing collisions is a direct measure of the vehicle's compliance to safety standards. Collision avoidance commonly relies on flexible reactions to hazardous areas. Once a hazard is detected, the system assesses the risk by analyzing the relative speed, distance, and trajectory of surrounding objects~\cite{9925071}. Additionally, some autonomous driving systems evaluate possible decisions to avoid collisions while maintaining high efficiency~\cite{botros2023spatio,10363676}. Furthermore, other autonomous driving systems use rule-based commands to adjust the AV's behavior when unsafe conditions emerge~\cite{bouchard2022rule}. For instance, AVs will be asked to stop when they encounter an interaction and spot-lines simultaneously~\cite{bouchard2022rule}.

\subsection{Driving Efficiency}
Driving efficiency refers to an AV’s ability to maintain a high average speed while adapting to varying traffic conditions. However, the implications of driving efficiency extend far beyond speed, affecting road capacity, user experience, and energy consumption.

On road capacity, efficient driving allows vehicles to travel at optimal speeds, minimizing delays and reducing traffic congestion. For example, HDVs tend to drive faster on familiar roads, contributing to higher road capacity and traffic flow~\cite{gajjar2016critical,kamal2018road,kamal2016efficient}. Similarly, AVs promote smoother traffic flow when they operate efficiently. Therefore, efficient driving allows more vehicles to travel smoothly without congestion.
On user experience, an efficient journey means shorter travel time and a smoother ride, significantly improving overall satisfaction~\cite{hensher2011valuation,steck2018autonomous,zhai2018ecological}.
 Besides, improving driving efficiency is crucial for reducing the energy consumption~\cite{birrell2014effect,sun2020optimal}.

\subsection{Training Efficiency}
Training efficiency of algorithms directly impact the time and resources required to bring a fully functional AV system to reality. 
One primary benefit of improved training efficiency is the reduced training time. The acceleration allows developers to focus more on system fine-tuning and extensive testing. Several studies have reduced training time by adding extra training mechanisms or adjusting the structures of networks~\cite{srinivas2021bottleneck,tian2024efficient,xu2021nn,touvron2022resmlp}. 
Another important benefit is the reduction in device wear and tear. Fast and efficient training reduces the required computational resources. By improving training efficiency, the workload of computing equipment is minimized, resulting in less frequent maintenance and replacement.

\subsection{Unselfishness}

In the context of autonomous driving, unselfishness refers to an AV's ability to consider and accommodate the intentions of other HDVs on the road. Unselfishness evaluates how well an AV can cooperate with surrounding vehicles by predicting their intentions and adjusting its behavior accordingly. Human drivers often prioritize factors such as safety, efficiency, and comfort, and these intentions vary widely depending on the specific situations.

Accurately classifying these driving intentions is essential for effective interactions with surrounding HDVs. Existing methods for recognizing driving intentions and enabling interaction-aware driving have been reviewed in \cite{martinez2017driving}. These methods categorize driving intentions across various scenarios, including car following and lane changing~\cite{li2018estimating,xu2018aware}. While many papers have focused on the self-driving characteristics of the ego vehicle \cite{jardin2020rule,vogel2022you}, the importance of unselfish behavior is becoming increasingly recognized.

An unselfish AV that effectively anticipates and responds to the intentions of other vehicles contributes to a smoother and more harmonious traffic flow. By avoiding overly aggressive or excessively cautious driving behaviors, the AV can help minimize disruptions and conflicts with other vehicles. This cooperative approach enhances the safety and efficiency of all vehicles in the road network and improves the driving experience for everyone involved.

\subsection{Algorithm Interpretability}
Algorithm interpretability has gained significant importance due to DRL models are required to make logical decisions. A logical structure makes the black-box of learning more transparent \cite{huang2020survey,fan2021interpretability}. In DRL-based autonomous driving systems, improving interpretability is crucial for system's safety and transparency. To address the challenges in interpretability, various approaches have been adopted, including policy visualization to showcase DRL behaviors \cite{9349146,ding2023saliendet}, and surrogate models for approximate human-understandable explanations \cite{9964561,gyevnar2023causal,dassanayake2021deep,zemni2023octet}. Furthermore, specific rule-based methods, algorithmic structure-adapted methods, and human-grounded methods have been proposed to assess interpretability.

Specific rules have been developed to assess interpretability \cite{chen2023interpretable}. One such rule, known as $FAST$, evaluates interpretability via four criteria: $F$ for fairness, $A$ for accountability, $S$ for sustainability, and $T$ for transparency \cite{leslie2019understanding}. Fairness requires models to be formalized using basic explanation labels and functionality evaluation. Accountability refers to answerability and auditability, ensuring that the system has been clearly defined. Sustainability ensures safe operation without inequality or discrimination, while transparency ensures that the model's internal rule settings are accessible and understandable.  

Some methods focuses on assessing interpretability by adjusting DRL algorithmic structures. Researchers achieve this by developing standardized benchmarks that use interpretability metrics \cite{hooker2019benchmark,tomsett2020sanity} or by troubleshooting explanations \cite{adebayo2018sanity,ghorbani2019interpretation} to identify instances where these explanations fall short. In addition, some studies concentrate on altering neural network architectures to enhance interpretability \cite{ismail2019input,wu2018beyond}.

Furthermore, human-grounded methods focus on how easily people can understand the model’s key computational sections \cite{speith2023new}. DRL-based algorithms incorporating traffic-related models enable people to better understand their structures through traffic knowledge or mathematical formulation, thereby improving interpretability.

\section{Deep Reinforcement Learning-based decision-making on Highways}

\subsection{Single-factor Methods for Highway Driving}
Many works consider only one of five key factors. A Double Deep Q-Network (DDQN) is integrated with handcrafted safety and dynamically-learned safety modules in~\cite{9304744}. The handcrafted safety module relies on heuristic safety rules derived from common driving practices, ensuring a $d_s$ with other vehicles. The dynamically-learned safety module uses driving data to learn safety patterns. By integrating both the handcrafted and dynamically-learned safety modules, the driving safety is improved. 

Moreover, deep deterministic policy gradients (DDPG) have been used to improve driving efficiency by overtaking surrounding vehicles in~\cite{kaushik2018overtaking}. The overtaking-oriented training is achieved by adding a high reward for overtaking maneuvers. The reward function for overtaking is formulated as~\cite{kaushik2018overtaking}:
\begin{equation}
    R_\textrm{overtaking}= R_\textrm{lane\_keeping} + 100 \times (n - \textrm{RacePos})
\end{equation}
where $R_\textrm{lane\_keeping}$ is the reward for lane-keeping, $n$ is the total number of vehicles in a given episode, and $RacePos$ reflects the number of vehicles in front of the AV. Therefore, the larger the $RacePos$, the smaller the $R_\textrm{overtaking}$. Although safety rewards are applied, the collision rate increases with the frequency of overtaking.

Furthermore, non-linear model predictive control (NMPC) has been integrated with DDQN to maintain safe highway driving in~\cite{albarella2023hybrid}. NMPC inherently incorporates vehicle dynamics as constraints into its optimization, ensuring that the control inputs from the DRL agent remain within safe and feasible bounds \cite{albarella2023hybrid}:
\begin{equation}
\begin{aligned}
    \min_{x(t),u(t)} & \int_0^T e(t)^\top Q e(t) + r \delta^2(t) + r u^2(t) \textrm{d}t \\
    \text{s.t.} \quad & \dot{x}(t) = f(x(t), u(t)), \\
    & e_{y_\textrm{min}} \leq e_y(t) \leq e_{y_\textrm{max}}, \\
    & e_{\psi_\textrm{min}} \leq e_\psi(t) \leq e_{\psi_\textrm{max}}, \\
    & \delta_\textrm{min} \leq \delta(t) \leq \delta_\textrm{max}, \\
    & u_{\textrm{min}} \leq u_1(t) \leq u_{\textrm{max}}
\end{aligned}
\end{equation}
where $T$ is the prediction horizon, $e(t)$ is the error vector to be regulated to zero, and $Q = \text{diag}(q_1, q_2)$ is a diagonal matrix of tracking weights. The control effort weight is denoted by $r$. The steering angle is represented by $\delta(t)$, and the control input is $u(t)$. The state vector is $x(t)$, and $f$ represents the system dynamics. $e_y(t)$ and $e_\psi(t)$ are the lateral position error and heading angle error, respectively. The variables $e_{y_\textrm{min}}$, $e_{y_\textrm{max}}$, $e_{\psi_\textrm{min}}$, $e_{\psi_\textrm{max}}$, $\delta_\textrm{min}$, $\delta_\textrm{max}$, $u_{\textrm{min}}$, and $u_{\textrm{max}}$ are the minimum and maximum admissible values for the lateral position error, heading angle error, steering angle, and control input, respectively. NMPC improves the interpretability of safe control by providing a clear mathematical formulation that integrates the system's constraints with the agent's decision-making~\cite{meng4876862optimizing,lin2024conflicts,gomez15explaining}.

Additionally, a policy gradient (PG) method has been used with hard constraints to ensure safe highway driving in~\cite{shalev2016safe}. These hard constraints prevent the AV from approaching risky boundaries, such as track edges. For example, the AV's longitudinal and lateral positions are restricted from approaching the track boundaries. Cooperative lane-changing has been achieved in~\cite{yu2019reinforcement}, enhancing the unselfishness. Interpretability has been improved by combining DRL with imitation learning (IL) in~\cite{10296611}. IL uses expert demonstrations to make the learning more interpretable. Training efficiency in highway driving is also enhanced by integrating a spatial attention module and attention mechanism into the deep Q-network in~\cite{9240387}.

\subsection{Dual-factor Methods for Highway Driving}
Additionally, two of the five considered factors are integrated in some recent studies. The Intelligent Driver Model (IDM)~\cite{treiber2000congested} has been incorporated into the DDQN for highway driving in~\cite{8914621}. The IDM prevents collisions during car-following and therefore, the integration of DDQN with IDM enhances both the driving safety and interpretability in highway driving. The IDM is formulated as~\cite{8914621}:
\begin{equation}
    U _{\textrm{IDM}} = U_{\textrm{max}} \left[ 1- \left(\frac{v_{\textrm{FV}}}{v_{e}} \right)^{4} - \left(\frac{g^{*}}{g} \right)^{2} \right]
    \label{eq33:idm}
\end{equation}
where $U_{\textrm{max}}$ is the maximum acceleration of the AV, $v_{e}$ is the expected velocity, and $g$ is the gap between the AV and the HDV. The desired gap $g^{*}$ between the AV and the front HDV is formulated as~\cite{8914621}:
\begin{equation}
    g ^{*} = d_s + v_{\textrm{AV}} T_{e} - \frac{v_{\textrm{AV}} \Delta v}{2 \sqrt{U_{\textrm{max}} b}}
\end{equation}
where $T_{e}$ is the expected time gap, $\Delta v$ is the velocity difference between the AV and the front vehicle (FV), and $b$ is the comfortable deceleration.

The reward function of DDQN has been adapted to improve driving safety and efficiency in~\cite{zhao2020deep}. Specifically, a penalty is applied when the vehicle goes off-road or the time-to-collision (TTC) falls below a threshold~\cite{zhao2024spatial} . The reward for driving efficiency is formulated as~\cite{zhao2020deep}:
\begin{equation}
    R = \frac{v_0}{v_\textrm{max}}
\end{equation}
where $v_\textrm{max}$ is the maximum velocity, and $v_0$ is the current velocity. This reward function helps maintain a relatively high driving velocity, thus increasing driving efficiency. Moreover, driving safety and altruism have been achieved using a level-$k$ game-based DQN in~\cite{albaba2021driver}. The level-$k$ game models the reasoning interaction between AVs and HDVs, promoting unselfish decision-making. A crash penalty is implemented in the DQN to prevent frequent collisions between AVs and HDVs. Additionally, unselfishness and training efficiency have been considered in~\cite{yang2018cm3}. Unselfishness is achieved through a cooperative multi-goal credit function-based policy gradient (PG). This adapted PG accounts for the goals of all vehicles, optimizing overall performance during training. Training efficiency is improved by a multi-agent reinforcement learning (MARL) curriculum, which reduces the number of trainable parameters and lowers computational costs. 

Unselfishness and driving efficiency on highways are achieved in~\cite{11}. Unselfishness is promoted through MARL by considering each vehicle's state. Driving efficiency is enhanced by a reward function that selects actions to increase the average velocity of all vehicles. Driving safety and driving efficiency have been achieved using multi-objective approximate policy iteration (MO-API) in~\cite{xu2018reinforcement}. Driving safety is ensured by monitoring collisions, while driving efficiency has been assessed by comparing the $v_0$ with the $v_{e}$. In~\cite{bai2019deep}, driving efficiency and unselfishness are considered in highway driving. Driving efficiency is achieved by a reward based on $v_0$, $v_\textrm{max}$, and $v_\textrm{min}$. Unselfishness is achieved by penalizing unnecessary lane changes to reduce disturbances to HDVs. Driving safety and training efficiency have been addressed in~\cite{10055747}. Safety is maintained by ensuring a $d_s$ between vehicles using rule-based constraints, while training efficiency is improved by incorporating a multi-head attention mechanism.

\subsection{Three-factor Methods for Highway Driving}
Furthermore, three of the five considered factors are combined in a few recent papers. Driving safety, interpretability, and driving efficiency have been improved in~\cite{liao2020decision}. Driving safety and interpretability are enhanced by using a collision penalty and the IDM. Driving efficiency is ensured by a reward based on the velocity difference between $v_\textrm{max}$ and $v_0$. The reward at time step $t$ is formulated as~\cite{liao2020decision}:
\begin{equation}
    R_t = -\text{Collision} - 0.1 \times (v_\textrm{max}^{t} - v_0^{t}) - 0.4 \times (L - 1)^2
\end{equation}
where $\text{Collision}$, $v_\textrm{max}^{t}$, and $v_0^{t}$ are the occurrence of a collision, maximum velocity, and current velocity at time $t$, respectively. $L$ represents the relative position the target lane, where $L=1$ indicates that the vehicle has successfully reached the target lane. A collision results in a negative reward, and a larger difference between $v_0$ and $v_\textrm{max}$ also leads to a negative reward. Additionally, if the vehicle does not drive in the target lane, a penalty is applied. 

A multi reward-based DQN has been proposed to achieve safe, efficient, and unselfish driving in~\cite{8917304}. Three rewards are combined: speed reward, limited lane-changing reward, and overtaking reward. The speed reward is a normalized reward based on the current speed relative to the minimum and maximum speed limits~\cite{8917304}:
\begin{equation}
    R_v = \frac{(v_0 - v_\textrm{min}) * r_v}{v_\textrm{max} - v_\textrm{min}}
\end{equation}
where $R_v$ represents the reward for speed, encouraging higher speeds within safe limits. $v_\textrm{min}$ is the minimum speed of the agent vehicle, and $r_v$ is the base reward for speed. The limited lane-changing reward function is designed to minimize the number of lane changes, promoting safer driving and reducing the disturbance to surrounding vehicles:
\begin{equation}
R_l =
\begin{cases}
-r_l, & \text{if the agent vehicle changes lanes;} \\
0, & \text{otherwise.}
\end{cases}
\end{equation}
where $-r_l$ is the penalty value for a lane change. The overtaking reward function encourages the agent vehicle to overtake more vehicles, improving driving efficiency:
\begin{equation}
R_o =
\begin{cases}
r_o, & \text{if the agent vehicle overtakes another vehicle;} \\
0, & \text{otherwise.}
\end{cases}
\end{equation}
where $r_o$ is the reward value for overtaking. 

Interpretability, driving safety, and driving efficiency have been achieved in~\cite{aradi2018policy}. Safety and efficiency are enhanced by penalizing frequent lane changes and tracking the desired velocity $v_d$, respectively. Interpretability is achieved through a car-following process using a proportional-derivative (PD) controller with transparent mathematical formulation~\cite{aradi2018policy}:
\begin{equation}
    d_{\textrm{des},i} = \alpha_i^j v_l^{j+1}
\end{equation}
\begin{equation}
    a_{\textrm{cf},i} = K_p(x_l^{j+1} - x_i^j) + K_d(v_l^{j+1} - v_i^j)
\end{equation}
where $d_{\textrm{des},i}$ is the desired following distance for the $i$-th vehicle, $\alpha_i^j$ is a sensitivity parameter with random values from $\mathcal{N}(1.3, 0.02)$, $v_l^{j+1}$ is the speed of the leading vehicle in the $(j+1)$-th lane, $a_{\textrm{cf},i}$ is the acceleration command, $K_p$ and $K_d$ are the proportional and derivative gains, and $x_l^{j+1}$ and $x_i^j$ are the positions of the leading and $i$-th vehicles, respectively.

In~\cite{wang2021tactical}, driving safety, efficiency, and interpretability have also been combined. Safety and efficiency are achieved by penalizing collisions and rewarding high average velocity. Interpretability is enhanced by using the risk potential field (RPF), which models and visualizes risks around surrounding vehicles. In~\cite{9972471}, driving safety and interpretability have been achieved in adaptive cruise control (ACC), which maintains $d_s$ between vehicles and provides interpretable mathematical formulations. Driving efficiency is achieved by rewarding each high-speed state. Finally, driving safety, driving efficiency, and training efficiency have been achieved in~\cite{9044113}. Safety is ensured through a collision penalty, and efficiency is rewarded based on the velocity difference between $v_0$ and $v_\textrm{min}$. Training efficiency is improved by using a long short-term memory (LSTM) network-assisted DDQN.

\subsection{Four-factor Methods for Highway Driving}
Moreover, four of the five considered factors have been included in some studies. Driving safety, driving efficiency, training efficiency, and interpretability have been considered in~\cite{lv2022safe}. Driving safety and driving efficiency are achieved by a reward function that maintains a $d_s$ from the leading vehicle while tracking the $v_d$. Interpretability is ensured through safety-based driving rules~\cite{lv2022safe}:
\begin{equation}
t_{f_\textrm{min}} = \inf\left\{t : t > \frac{2(v - v_{f_\textrm{target}})}{a_{d_\textrm{max}}}\right\}
\end{equation}
\begin{equation}
t_{b_\textrm{min}} = \inf\left\{t : t > \frac{2(v_{b_\textrm{target}} - v)}{a_{d_\textrm{max}}}\right\}
\end{equation}
\begin{equation}
d_{\text{target}_\textrm{min}} = \min \left\{ \frac{(v - v_f) t_f}{2}, \frac{(v_b - v) t_b}{2} \right\}
\end{equation}
\begin{equation}
\Delta d_\textrm{target} = \min\left\{|x_\text{AV} - x_{f_\textrm{target}}|, |x_\text{AV} - x_{b_\textrm{target}}|\right\}
\end{equation}
where $t_{f_\textrm{min}}$ and $t_{b_\textrm{min}}$ are the minimum safe time intervals between the AV and the vehicles in front and behind in the target lane, respectively. $v$ is the speed of the AV; $v_{f_\textrm{target}}$ and $v_{b_\textrm{target}}$ are the speeds of the front and behind vehicles in the target lane, respectively. $a_{d_\text{max}}$ is the maximum deceleration. $d_{\text{target}_\textrm{min}}$ is the minimum distance between the AV and the FV in the target lane, and $\Delta d_\textrm{target}$ is the actual distance between the AV and the nearest vehicle in the target lane. $x_\text{AV}$, $x_{f_\textrm{target}}$, and $x_{b_\text{target}}$ represent the horizontal coordinates of the AV, the front target vehicle, and the vehicle behind in the target lane, respectively. By implementing these safety rules, the decision-making of the AV becomes more transparent and interpretable. Training efficiency is achieved through the potential-based reward shaping function. The total reward function and reward shaping function are given by~\cite{lv2022safe}:
\begin{equation}
R' = R(s, a, s') + \beta F(s, a, s')
\end{equation}
\begin{equation}
F(s, a, s') = \gamma \phi(s') - \phi(s)
\end{equation}
\begin{equation}
F(s, a, s', t, t') = \gamma \phi(s', t') - \phi(s, t)
\end{equation}
where $R'$ is the new reward criterion, $R(s, a, s')$ is the original reward function, $\beta$ is a weighting factor, $F(s, a, s')$ is the potential-based reward shaping function, $s$ and $s'$ are the current and next state, respectively, $a$ is the action taken, and $\gamma$ is the discount factor. $\phi(s)$ is the potential function mapping the state to a real number, and $t$ and $t'$ are the time corresponding to $s$ and $s'$, respectively.  (16) combines the original reward function with an additional shaping term. (17) defines the shaping function as the difference between the discounted potential of the next state and the current state. (18) extends (17) by including time as a parameter and therefore allows for dynamic potential functions. 

Driving safety, driving efficiency, unselfishness, and training efficiency on highways have been addressed in~\cite{ruadulescu2019deep}. Driving safety and efficiency are considered in the reward function of the DQN. Unselfishness is achieved through a joint policy update, accounting for the profits of multiple vehicles. Training efficiency is enhanced by reusing the experiences of single agents within a MARL framework. In~\cite{8638814}, driving safety, efficiency, unselfishness, and training efficiency on highways have been explored. Safety and efficiency are ensured by assessing the remaining reaction time during emergencies and selecting the proper lane-changing point, respectively. Unselfishness is achieved using MARL for cooperative highway driving, while training efficiency is improved with a dynamic coordinate graph (DCG) that enhances cooperative efficiency. In~\cite{kaushik2019parameter}, safety, efficiency, unselfishness, and training efficiency have been considered. Safety is ensured by applying penalties both for collisions and for deviating from the road. Efficiency is achieved by rewarding each state that overtakes other vehicles. Unselfishness is promoted through MARL to coordinate driving. Training efficiency is enhanced by employing a parameter-sharing mechanism, which stores experience of each agent to reinforce common scenario understanding. 

In~\cite{9483545}, safety, efficiency, unselfishness, and interpretability have been considered. Safety, efficiency, and unselfishness are improved through rewards for collisions, velocity ratio between $v_0$, $v_\textrm{max}$, and $v_\textrm{min}$, and limiting unnecessary lane changes, respectively. Interpretability is enhanced by integrating an autonomous emergency braking system, promoting safer decision-making. In~\cite{10214640}, safety, efficiency, interpretability, and training efficiency have been addressed. Safety and efficiency are enhanced by adding a safety layer and incorporating the ratio between longitudinal speed, $v_\textrm{max}$, and $v_\textrm{min}$. Interpretability is improved by using a support vector machine (SVM), which provides interpretable safe decision boundaries. Training efficiency is boosted through an external space attention mechanism that pays attention to the crucial areas of surrounding environment. 

In~\cite{8723635}, safety, efficiency, unselfishness, a

nd training efficiency have been tackled. Safety, efficiency, and unselfishness are achieved through rewards for collisions, velocity ratios, and MARL, while training efficiency is improved using a distributional DQN with multi-type input data. Finally, in~\cite{9497870}, safety, efficiency, unselfishness, and interpretability have been considered. Safety, efficiency, and unselfishness are enhanced through rewards for collisions, target velocity differences, and unnecessary lane changes, respectively. Interpretability is achieved through rule-based constraints, such as preventing lane changes with short lateral distances to lead vehicles.

\subsection{Five-factor Methods for Highway Driving}
Additionally, all the five factors are addressed in some studies, such as~\cite{9857596}. Driving safety and efficiency, and unselfishness are achieved by reducing collisions, increasing speed, and minimizing lane-change frequency through rewards. Training efficiency is improved through a convolutional neural network-based LSTM. Interpretability is enhanced by using spatio-temporal image representations for HDVs, which increase the interpretability of the inputs. The DRL-based decision making in highway driving based on DDTUI is summarized in Table I.

\begin{table*}[t]
\centering
\captionsetup{labelfont={sc}, textfont={sc}, labelsep=newline, justification=centering}
\caption{Evaluation of the DRL-based decision making in highway driving}
\label{tab:comprehensive_highway_driving_factors}
\vspace{-2mm}
\setlength{\tabcolsep}{8pt}
\renewcommand{\arraystretch}{1.53}
\small
\begin{tabular}{lccccc}
\hline
Reference & Safety & Efficiency & Training Efficiency & Unselfishness & Interpretability \\
\hline
\cite{9304744} & Safety modules & - & - & - & - \\
\cite{kaushik2018overtaking} & - & Overtaking reward & - & - & - \\
\cite{albarella2023hybrid} & NMPC constraints & - & - & - & - \\
\cite{shalev2016safe} & Hard constraints & - & - & - & - \\
\cite{yu2019reinforcement} & - & - & - & Local interactions & - \\
\cite{10296611} & - & - & - & - & Imitation learning \\
\cite{9240387} & - & - & Attention module & - & - \\
\cite{8914621} & IDM integration & - & - & - & IDM integration \\
\cite{zhao2020deep} & TTC threshold & Velocity reward & - & - & - \\
\cite{albaba2021driver} & Crash penalty & - & - & Level-k game & - \\
\cite{yang2018cm3} & - & - & MARL curriculum & Cooperative function & - \\
\cite{11} & - & Average velocity & - & MARL & - \\
\cite{xu2018reinforcement} & Collision monitoring & Velocity comparison & - & - & - \\
\cite{bai2019deep} & - & Velocity reward & - & Lane change penalty & - \\
\cite{10055747} & Rule-based & - & Attention mechanism & - & - \\
\cite{liao2020decision} & IDM \& collision & Velocity difference & - & - & IDM integration \\
\cite{8917304} & Speed-limit reward & Overtaking reward & - & Lane-change limit & - \\
\cite{aradi2018policy} & Lane change penalty & Velocity tracking & - & - & PD controller \\
\cite{wang2021tactical} & Reward function & Reward function & - & - & Risk potential field \\
\cite{9972471} & Adaptive cruise & High-speed reward & - & - & ACC formulations \\
\cite{9044113} & Collision penalty & Velocity difference & LSTM-DDQN & - & - \\
\cite{lv2022safe} & Safety rules & Reward function & Reward shaping & - & Safety rules \\
\cite{ruadulescu2019deep} & Reward function & Reward function & MARL reuse & Joint policy & - \\
\cite{8638814} & Reaction time & Lane-changing point & DCG efficiency & MARL & - \\
\cite{kaushik2019parameter} & Collision penalties & Overtaking reward & Parameter sharing & MARL & - \\
\cite{9483545} & Collision rewards & Velocity ratio & - & Lane change limit & Emergency braking \\
\cite{10214640} & Safety layer & Velocity ratio & Attention mechanism & - & SVM boundaries \\
\cite{8723635} & Collision rewards & Velocity ratio & Distributional DQN & MARL & - \\
\cite{9497870} & Collision rewards & Velocity difference & - & Lane change penalty & Rule-based \\
\cite{9857596} & Collision reduction & Speed increase & CNN-LSTM & Lane change limit & Representations \\
\hline
\end{tabular}
\begin{tablenotes}
\footnotesize
\item '-' indicates that the corresponding factor was not explicitly addressed in the study.
\end{tablenotes}
\end{table*}

\section{Deep Reinforcement Learning-based decision-making in On-ramping Merging}
\begin{table*}[t]
\centering
\captionsetup{labelfont={sc}, textfont={sc}, labelsep=newline, justification=centering}
\caption{Evaluation of the DRL-based decision making in on-ramping merging}
\label{tab:ramp_driving_factors_summary}
\vspace{-2mm}
\setlength{\tabcolsep}{19pt}
\renewcommand{\arraystretch}{1.6}
\small
\begin{tabular}{@{}p{1.2cm}p{2cm}p{2cm}p{2cm}p{2cm}p{2cm}@{}}
\hline
Ref. & Safety & Efficiency & Training Efficiency & Unselfishness & Interpretability \\
\hline
\cite{liu2021deep} & - & Reward function & - & - & - \\
\cite{yang2019deep} & - & Travel time reward & - & - & - \\
\cite{8317735} & Safety factor & - & - & - & - \\
\cite{lin2020anti} & Collision-free driving & - & - & - & - \\
\cite{8916903} & - & Average velocity reward & - & MARL & - \\
\cite{cheng2022adaptive} & - & Error state reduction & - & - & Traditional controller \\
\cite{9922173} & \multirow{2}{*}{Distance penalty} & \multirow{2}{*}{Distance minimization} & \multirow{2}{*}{-} & \multirow{2}{*}{MARL} & \multirow{2}{*}{-} \\
 & & & & & \\
\cite{hu2024guided} & - & Trip time difference & Teacher-student model & - & Traditional control \\
\cite{deng2023automated} & - & Speed comparison & - & Ramp metering & Ramp metering \\
\cite{wang2022integrated} & - & DDPG-assisted RM & DDPG & RM and VSL & RM and VSL \\
\cite{10423843} & APF & MPC with DDQN & MPC with DDQN & - & APF \\
\cite{9557770} & \multirow{2}{*}{Collision penalty} & Stop maneuver & \multirow{2}{*}{DIM with DDPG} & HDV & \multirow{2}{*}{-} \\
 & & penalty & & intentions & \\
\cite{10215357} & Safety reward & Efficiency reward & IPPO & - & IDM \\
\cite{10159552} & \multirow{2}{*}{Crash evaluation} & Stable speed & Safety & \multirow{2}{*}{MARL} & Rule-based \\
 & & assessment & supervisor & & constraints \\
\cite{9994638} & Collision rewards & Velocity ratio & Adversarial constraints & Nash-based game & transparent game process \\
\cite{9954271} & DRAC & Velocity ratio reward & Multi-state rep. & Vehicle coop. & DRAC \\
\hline
\end{tabular}
\end{table*}

\subsection{Single-factor Methods for On-ramping Merging}
Driving efficiency has been considered using Q-learning in~\cite{liu2021deep}. The remaining time of AV on the ramp lane is reduced by optimizing the reward function, thus promoting fast lane-changing to the main lane. The reward function is formulated as~\cite{liu2021deep}:
\begin{equation}
r_t = \mu \bar{v}_t + \omega \bar{q}_t, \quad \mu > 0, \quad \omega < 0
\end{equation}
where \( r_t \) represents the reward after taking action \( a_t \); \( \bar{v}_t \) denotes the average speed in the merging area during time step \( t \); \( \bar{q}_t \) indicates the average queue length at the on-ramp during time steps \( t \) and \( t + 1 \); \( \mu \) is a positive weight assigned to the speed reward, and \( \omega \) is a negative weight for the queue length reward. These rewards help balance the trade-off between enhancing vehicle mobility on the mainline and reducing delays at the on-ramp. Driving efficiency has also been improved in~\cite{yang2019deep} by reducing the total travel time reward (\(R_\textrm{TTT}\)), represented by the summation of the total number of vehicles at each time step. Driving safety has been achieved through a safety factor in~\cite{8317735}. The safety factor is a negative reward when the relative distances between AV and HDV are small. Driving safety has been achieved in~\cite{lin2020anti}, by giving rewards for each state having a $d_s$ and penalties for collisions.

\subsection{Dual-factor Methods for On-ramping Merging}
Driving efficiency and unselfishness have been considered in~\cite{8916903}. Driving efficiency is achieved by using the average velocity of AVs as part of the reward, and unselfishness is achieved using MARL to maximize general profits. Interpretability and driving efficiency have been addressed in~\cite{cheng2022adaptive}. 
Interpretability is achieved by using DDPG to tune a traditional controller's parameters, keeping the traditional controller as the main system to ensure transparency. Driving efficiency is enhanced by reducing the error state, which reflects the gap between actual and critical traffic density. A smaller error state leads to higher traffic flow and average speed.

\subsection{Three-factor Methods for On-ramping Merging}
Driving efficiency, interpretability, and training efficiency have been addressed in~\cite{hu2024guided}. Driving efficiency is achieved through a reward using the difference between the start and end time of each trip. Training efficiency is improved by a teacher-student model to train the decision-making system, where the traditional control method acts as the teacher guiding the DQN student. Similarly, driving efficiency, interpretability, and unselfishness have been improved in~\cite{deng2023automated}. Driving efficiency is achieved by a reward that compares the average speed between two consecutive time. Unselfishness and interpretability are achieved by combining ramp metering (RM) with Q-learning. RM optimizes average vehicle speed and is algorithmically transparent.  Driving safety, efficiency, and unselfishness have been addressed in~\cite{9922173}. Driving safety is achieved through a penalty for small relative distances, driving efficiency is enhanced by minimizing the relative distance while maintaining at least the safe distance, and unselfishness is achieved using MARL to optimize general driving performance.
\subsection{Four-factor Methods for On-ramping Merging}
Driving efficiency, training efficiency, unselfishness, and interpretability have been improved in~\cite{wang2022integrated}, where driving and training efficiency is enhanced by DDPG-assisted RM and variable speed limit (VSL) control. Interpretability and unselfishness are improved through RM and VSL, which are algorithmically transparent. Driving safety, efficiency, training efficiency, and interpretability have been achieved in~\cite{10423843}, where safety and interpretability are enhanced by combining APF, which quantifies and visualizes risk areas and provides interpretable input. Driving and training efficiency are achieved by combining MPC with DDQN, which outperforms single MPC or DDQN methods. Similarly, driving safety, efficiency, training efficiency, and unselfishness have been addressed in~\cite{9557770}, where safety and efficiency are promoted by penalties for collisions and stop maneuvers. Training efficiency is improved by integrating the driver’s intention model (DIM) with DDPG, while unselfishness is achieved by considering HDVs' various cooperation intentions. In~\cite{10215357}, driving safety, efficiency, training efficiency, and interpretability have been achieved by applying the safety, efficiency rewards, and IDM respectively, with independant PPO (IPPO) used for improved training efficiency compared to baseline algorithms.

\subsection{Five-factor Methods for On-ramping Merging}
In~\cite{10159552}, driving safety and efficiency have been achieved through collision and stable speed assessment rewards, respectively. Training efficiency is improved by using a safety supervisor, filtering detectable collision cases. Interpretability is enhanced through rule-based safety constraints, and unselfishness is achieved using MARL to maximize general profits. Similarly, all factors have been addressed in~\cite{9994638}, where driving safety and efficiency are achieved via collision rewards and a velocity ratio, respectively. Training efficiency is enhanced by adversarial constraints, while unselfishness and interpretability is enhanced through a transparent Nash-based game that considers HDV's profits. Finally, in~\cite{9954271}, driving safety and interpretability have been achieved using the deceleration rate to avoid a crash (DRAC), which has a detailed mathematical formulation and is transparent. Driving efficiency is improved by using (5) as an efficiency reward. Unselfishness is addressed by considering the cooperation intentions of other vehicles, and training efficiency is improved using multi-state representations to enhance the agent's learning capabilities. The DRL-based decision making on on-ramp merging based on DDTUI is summarized in Table II.

\section{Deep Reinforcement Learning-based decision-making at Roundabouts}
\subsection{Single-factor Methods for Roundabout Driving}
Driving efficiency in roundabout driving has been improved using soft actor-critic (SAC) with higher peak rewards in~\cite{chen2019model}. Training efficiency has been achieved through action repeat and asynchronous advantage in~\cite{bacchiani2019microscopic}. Action repeat improves efficiency by allowing the agent to repeat the same action for several time steps, decreasing the frequency of making new decisions. Asynchronous advantage enables each agent to share its interaction experience with others. Training efficiency has been further improved by embedding the operational design domain (ODD) into DQN in~\cite{10667129}. ODD guides the training to more targeted scenarios, reducing unnecessary exploration and accelerating convergence.

\subsection{Dual-factor Methods for Roundabout Driving}
Driving efficiency and training efficiency are improved using the Conditional Representation Model (CRM) in~\cite{tian2020variational}, which helps the agent better understand safety by defining each state as safe or unsafe state. Training efficiency and interpretability have been improved by leveraging labeled data from domain experts as guidance in~\cite{wang2022imitation}. Driving safety and driving efficiency have been enhanced in~\cite{ferrarotti2024autonomous} by incorporating $v_d$ and allowable relative distance into the reward function. 

Training efficiency and interpretability have been improved in~\cite{9311168}. Training efficiency is achieved through optimization-embedded DRL for adaptive decision-making, and interpretability is enhanced by transparent model-based optimization. Driving safety and unselfishness have been achieved in~\cite{9565031}, with safety ensured by penalizing collisions and ramping off the road, and unselfishness promoted by using MARL to maximize collective benefits. Driving safety and interpretability have been achieved in~\cite{10354212}, with safety maintained through penalties for collisions with HDVs and walls. Interpretability is supported by gradual training mode, similar to human learning, where the system starts with sparse traffic and progresses to dense traffic later.

\subsection{Three-factor Methods for Roundabout Driving}
Driving safety, driving efficiency, and training efficiency have been improved in~\cite{yuan2023safe}. Driving safety and driving efficiency are promoted through safety and efficiency rewards, respectively. Training efficiency is enhanced via trust region policy optimization (TPRO), which converges faster than PPO and DDPG. In~\cite{10698974}, driving safety and driving efficiency have been achieved by rewards for non-collision lane-changing and the difference between initial and target velocities, respectively. Training efficiency is improved by embedding LSTM into the actor-critic network. Training efficiency, driving safety and efficiency have been enhanced in~\cite{capasso2020simulation}. Training efficiency is improved by normalizing the initial reward for faster convergence. Driving efficiency and driving safety benefit from multiple environments where agents are trained simultaneously, achieving higher success rates and fewer crashes.

\subsection{Four-factor Methods for Roundabout Driving}
Driving safety, driving efficiency, training efficiency, and unselfishness have been addressed in~\cite{capasso2020intelligent}, where safety is maintained using $d_s$, and driving efficiency is enhanced by the ratio of initial to target velocity. Training efficiency is improved through a synthetic representation mechanism that enhances agents' understanding, and unselfishness is promoted using MARL to maximize joint benefits. Driving safety, driving efficiency, interpretability, and training efficiency have been addressed in~\cite{wang2016driving}, where safety is ensured via crash penalties and efficiency via high-speed rewards. Interpretability is maintained using the IDM for safe, transparent algorithmic-following. Training efficiency is improved through an interval prediction model to precompute feasible paths, reducing training computation. Driving safety, driving efficiency, training efficiency, and interpretability have been enhanced in~\cite{9757810}. Safety and efficiency are promoted through penalties for collisions and vehicle-stop maneuvers, respectively. Training efficiency is increased by integrating DDPG, DQN, and NMPC. Interpretability is enhanced via the NMPC.
\begin{table*}[t]
\centering
\captionsetup{labelfont={sc}, textfont={sc}, labelsep=newline, justification=centering}
\caption{Evaluation of the DRL-based decision making at roundabouts}
\label{tab:ramp_driving_factors_summary}
\vspace{-2mm}
\setlength{\tabcolsep}{19pt}
\renewcommand{\arraystretch}{1.55}
\small
\begin{tabular}{@{}p{1.2cm}p{2cm}p{2cm}p{2cm}p{2cm}p{2cm}@{}}
\hline
Ref. & Safety & Efficiency & Training Efficiency & Unselfishness & Interpretability \\
\hline
\cite{chen2019model} & - & SAC with higher peak rewards & - & - & - \\
\cite{bacchiani2019microscopic} & - & - & Action repeat, asynchronous advantage & - & - \\
\cite{10667129} & - & - & ODD-embedded DQN & - & - \\
\cite{tian2020variational} & - & CRM & CRM & - & - \\
\cite{wang2022imitation} & - & - & Expert guidance & - & Expert guidance \\
\cite{ferrarotti2024autonomous} & Allowable relative distance & $v_d$ & - & - & - \\
\cite{9311168} & - & - & Optimization-embedded DRL & - & Model-based optimization \\
\cite{9565031} & Collision penalties & - & - & MARL & - \\
\cite{10354212} & Collision penalties & - & - & - & Gradual training \\
\cite{yuan2023safe} & Safety rewards & Efficiency rewards & TPRO & - & - \\
\cite{10698974} & Non-collision rewards & Velocity difference rewards & LSTM-embedded actor-critic & - & - \\
\cite{capasso2020simulation} & Fewer crashes & Higher success rates & Reward normalization & - & - \\
\cite{capasso2020intelligent} & Safety distance & Velocity ratio & Synthetic representation & MARL & - \\
\cite{wang2016driving} & Crash penalties & High-speed rewards & Interval prediction & - & IDM \\
\cite{9757810} & Collision penalties & Vehicle-stop penalties & DDPG, DQN, NMPC integration & - & NMPC \\
\cite{lin2024conflicts} & Rule-based inspector & High-speed rewards & KAN-DQN & Rule-based planning & Rule-based inspector \\
\hline
\end{tabular}
\end{table*}
\begin{table*}[h]
\centering
\captionsetup{labelfont={sc}, textfont={sc}, labelsep=newline, justification=centering}
\caption{Evaluation of the DRL-based decision making at unsignalized intersections}
\label{tab:ramp_driving_factors_summary}
\vspace{-2mm}
\setlength{\tabcolsep}{19pt}
\renewcommand{\arraystretch}{1.5}
\small
\begin{tabular}{@{}p{1.2cm}p{2cm}p{2cm}p{2cm}p{2cm}p{2cm}@{}}
\hline
Ref. & Safety & Efficiency & Training Efficiency & Unselfishness & Interpretability \\
\hline
\cite{peng2021connected} & - & Velocity difference reward & - & - & - \\
\cite{pozzi2020ecological} & - & Time penalty & - & - & - \\
\cite{9385927} & - & Total waiting time & Background removal ResNet & - & - \\
\cite{quang2020proximal} & - & Velocity difference reward & - & - & IDM \\
\cite{bai2022hybrid} & - & Velocity-based reward & - & - & Safety-based rule policy \\
\cite{bautista2022autonomous} & - & Safe distance reward & - & - & MPC with TD3 \\
\cite{li2023coor} & - & Velocity ratio reward & - & - & Gridded coordination zone \\
\cite{9304542} & - & Goal attainment reward & DQN with sub-tasks & - & - \\
\cite{10440155} & - & - & Incentive communication & MARL & - \\
\cite{10120647} & - & D2-TSP & DDQN & - & - \\
\cite{10074992} & - & CIM-enhanced DQN & CIM-enhanced DQN & - & - \\
\cite{li2020adaptive} & - & Low-speed penalty & Multi-agent DQN & MARL & - \\
\cite{shu2021driving} & - & High-velocity reward & DQL with transfer learning & - & IDM \\
\cite{9564720} & Collision penalties & High-velocity reward & SAC with attention & - & - \\
\cite{hoel2020reinforcement} & Collision penalties & Goal attainment reward & RPF & - & - \\
\cite{antonio2022multi} & AIM & Constant time penalty & AIM and LSTM & MARL & - \\
\cite{10417752} & Collision penalties & Goal attainment reward & Mix-Attention Network & - & IDM \\
\cite{9939107} & Collision penalties & Low-velocity penalty & VD-MADQL & MARL & IDM \\
\hline
\end{tabular}
\end{table*}

\begin{table*}[h]
\centering
\captionsetup{labelfont={sc}, textfont={sc}, labelsep=newline, justification=centering}
\caption{Occurrence and Ratio of Evaluation Factors Across Different Scenarios}
\label{tab:factor_ratio}
\vspace{-2mm}
\setlength{\tabcolsep}{3.5pt}
\renewcommand{\arraystretch}{1.5}
\small
\begin{tabular}{@{}p{3cm}p{2.5cm}p{2.5cm}p{2.5cm}p{2.5cm}p{2.5cm}@{}}
\hline
Scenario & Safety & Efficiency & Training Efficiency & Unselfishness & Interpretability \\
\hline
Highway     & 23 (76.7\%) & 20 (66.7\%) & 11 (36.7\%) & 13 (43.3\%) & 11 (36.7\%) \\
Ramp        & 9 (56.25\%) & 14 (87.5\%) & 8 (50\%) & 8 (50\%) & 9 (56.25\%) \\
Roundabout  & 10 (62.5\%) & 11 (68.75\%) & 12 (75\%) & 3 (18.75\%) & 6 (37.5\%) \\
Intersection & 5 (27.7\%) & 17 (94.4\%) & 12 (66.7\%) & 4 (22.2\%) & 6 (33.3\%) \\
Total     & 47 (58\%) & 63 (77.8\%) & 44 (54.3\%) & 29 (35.8\%) & 32 (39.5\%) \\
\hline
\end{tabular}
\begin{tablenotes}
\footnotesize
\item The numbers in parentheses indicate the percentage of the total studies for each factor.
\end{tablenotes}
\end{table*}
\subsection{Five-factor Methods for Roundabout Driving}
All five factors have been considered in~\cite{lin2024conflicts}, where driving safety and interpretability are ensured by a rule-based action inspector. Driving efficiency is enhanced via high-speed rewards. Training efficiency is achieved through a Kolmogorov-Arnold network-enhanced DQN. Unselfishness is promoted through rule-based route planning that considers the varying distributions of HDVs on the roundabout. The DRL-based decision making in roundabouts based on DDTUI is summarized in Table III.

\section{Deep Reinforcement Learning-based decision-making at Unsignalized Intersections}

\subsection{Single-factor Methods for Intersection Driving}
Traffic efficiency has been improved by using the difference between $v_0$ and $v_d$ as a reward in~\cite{peng2021connected}. Additionally, a penalty is applied when the velocity drops below a threshold, further boosting traffic efficiency. In~\cite{pozzi2020ecological}, driving efficiency has been achieved by applying a constant penalty as long as the AV has not reached the target exits.

\subsection{Dual-factor Methods for Unsignalized Intersection Driving}
Both driving and training efficiency have been improved in~\cite{9385927}, where the driving efficiency is enhanced by using total waiting time (TWT) as part of the reward. Training efficiency is increased by employing a background removal ResNet as the Q-network, resulting in lower TWT than baseline algorithms. In~\cite{quang2020proximal}, driving efficiency and interpretability have been enhanced. The driving efficiency is improved by using the difference between the $v_d$ and $v_0$ as part of the reward, while the interpretability is achieved through the use of IDM for safe and transparent vehicle following. Similarly, in~\cite{bai2022hybrid}, both driving efficiency and interpretability have been improved. The former is enhanced by incorporating a velocity-based reward, and the latter is enhanced by applying a safety-based rule policy. In~\cite{bautista2022autonomous}, driving efficiency is increased by using the safe distance as a reward and the risky distance as a penalty, resulting in higher success rates. Interpretability is achieved using a model-based transparent method combined with twin delayed deep deterministic policy gradient (TD3). In~\cite{li2023coor}, driving efficiency is enhanced by the ratio of $v_0$ to $v_\textrm{max}$ as part of the reward, and interpretability is improved by gridding the coordination zone into different granularities, converting risky areas into a matrix format.

Both driving efficiency and training efficiency have been improved in~\cite{9304542}. Driving efficiency is achieved by rewarding goal attainment, and training efficiency is increased by using DQN with common and specific sub-tasks. The common sub-task enables knowledge sharing across tasks, while the specific sub-task helps the system better understand a task’s main goal. Training efficiency and unselfishness have been improved in~\cite{10440155} through an incentive communication-assisted MARL. Agents create custom messages to influence other agents’ policies, improving coordination and achieving globally optimal decisions. The unselfishness is realized by using MARL to maximize overall profits. In~\cite{10120647}, driving efficiency and training efficiency have been improved by the adaptive dual-objective transit signal priority (D2-TSP) algorithm with DDQN. D2-TSP optimizes bus speed, saving time for both passengers and those waiting at downstream stops. Similarly, in~\cite{10074992}, cooperative intersection management-enhanced DQN boosts both driving and training efficiency by leveraging connectivity between vehicles.

\subsection{Three-factor Methods for Unsignalized Intersection Driving}
In~\cite{li2020adaptive}, driving efficiency, unselfishness, and training efficiency have been addressed. Driving efficiency is enhanced by penalizing each low-speed state, while unselfishness is achieved through MARL for maximizing overall profits. Training efficiency is improved with multi-agent DQN, which offers faster convergence than baseline algorithms. In~\cite{shu2021driving}, driving efficiency, training efficiency, and interpretability have been integrated. Driving efficiency is increased by rewarding each high-velocity state, and training efficiency is achieved by combining deep Q-learning with transfer learning. Interpretability is improved by using the IDM for safe vehicle following.

In~\cite{9564720}, driving safety, driving efficiency, and training efficiency have been incorporated. Driving safety is promoted through collision penalties, and driving efficiency is enhanced by rewarding velocities higher than a baseline. Training efficiency is improved by using a spatial and temporal attention module with SAC. In~\cite{hoel2020reinforcement}, all three aspects have also been addressed. Driving safety and efficiency are improved by rewarding goal attainment and penalizing collisions. Training efficiency is enhanced using a randomized prior function (RPF) for each ensemble member, leading to a better Bayesian posterior~\cite{osband2018randomized}. 

\subsection{Four-factor Methods for Unsignalized Intersection Driving}
In~\cite{antonio2022multi}, driving safety, driving efficiency, training efficiency, and unselfishness have been incorporated. Driving safety is enhanced through autonomous intersection management (AIM), and driving efficiency is improved by applying a constant penalty until the AV reaches the exits. Training efficiency is improved by embedding AIM and LSTM into the learning, and unselfishness is achieved through MARL. In~\cite{10417752}, driving safety, driving efficiency, training efficiency, and interpretability have been integrated. Driving safety is promoted through collision penalties, and driving efficiency is enhanced by rewarding goal attainment. Training efficiency is improved through the Mix-Attention Network, synthetic representation mechanism, and replay memory mechanism. The interpretability is ensured by using the IDM.

\subsection{Five-factor Methods for Intersection Driving}
In~\cite{9939107}, driving safety has been promoted through collision penalties, while driving efficiency is enhanced by penalizing each state with velocity lower than the $v_\textrm{min}$. Training efficiency is improved using value decomposition-based multi-agent deep Q-learning. Unselfishness is achieved by employing MARL to minimize joint profits, and interpretability is ensured through the IDM. The DRL-based decision making on unsignalized intersections based on DDTUI is summarized in Table IV.
 
\section{Conclusion and Discussion}
This survey presents a comprehensive overview of the current state of the art in DRL-based decision-making for autonomous vehicles. By discussing recent research efforts in this field, this survey highlights the diverse algorithms developed to address decision-making tasks across various scenarios, including highways, on-ramping merging, roundabouts, and unsignalized intersections. Our analysis goes beyond simply presenting these algorithms by uncovering valuable insights, identifying key gaps in the current research, and highlighting emerging trends in DRL-based decision making for autonomous driving. While driving efficiency and safety are addressed across most studies, there is a growing trend towards addressing multiple DDTUI factors concurrently. Emerging approaches, such as MARL and the integration of traditional control methods with DRL, show promise in tackling complex challenges with increased unselfishness and interpretability in autonomous driving. 

Based on existing studies, Table V summarizes the distribution of evaluation factors considered in four typical scenarios. Most studies, i.e., 18 studies that account for 94.4\% of existing studies, prioritize efficiency at intersections to optimize travel time in complex and interaction-heavy environments. Efficiency is also emphasized at ramps in 14 studies (i.e., 87.5\%) to reduce congestion and streamline traffic flow. Safety is particularly emphasized on highways in 23 studies (i.e., 76.7\%), which addresses the importance of accident prevention in high-speed settings. In contrast, intersections address safety less often. Training efficiency is significant at roundabouts in 12 studies (i.e., 75\%) and unsignalized intersections in 12 studies (i.e., 66.7\%). This reflects a need for effective training methods to ensure smooth vehicle maneuvering in these challenging contexts. Interpretability is particularly valued at ramps in 9 studies (i.e., 56.25\%) and on highways in 11 studies (i.e., 36.7\%), respectively. This emphasizes understandable decision-making in these areas. Unselfishness receives less emphasis overall, although highways and ramps give it much attention. Future challenges are summarized as

\begin{enumerate}
    \item Achieving a balance between all five DDTUI factors in a single framework: This survey reveals that while many studies addressed multiple DDTUI factors, very few managed to incorporate all five factors simultaneously. For instance, only 3 out of 16 studies in roundabout scenarios and 1 out of 19 studies in intersection scenarios addressed all five factors. This highlights the complexity of developing a unified framework that can effectively balance DDTUI. Future research should focus on developing integrated frameworks that can holistically address five DDTUI factors concurrently.

    \item Improving the interpretability of DRL models without sacrificing performance: While some studies have made strides in improving interpretability, such as using IDM for interpretable car-following, many high-performing DRL models remain black boxes. Out of the reviewed papers, less than 40\% explicitly addressed interpretability. Furthermore, most papers considering interpretability use only one method. In the future, multiple interpretability methods can be applied to enhance interpretability, such as using the APF and IDM concurrently.

    \item Enhancing the unselfishness of AVs in complex, multi-agent environments: While approximately 50\% of studies use MARL to promote unselfishness, the complexity of real-world traffic scenarios presents uncertainties of driving behaviors. Future research should explore more sophisticated MARL techniques based on real-world experience. For example, combining game theory with driving style classification based on real-world datasets can better model the behaviors of HDVs.
\end{enumerate}

\footnotesize\bibliographystyle{IEEEtran}
\bibliography{IEEEabrv,zq_lib}

\end{document}